\documentclass{article}

\PassOptionsToPackage{numbers, compress}{natbib}

\usepackage[preprint]{neurips_2021}

\makeatletter
\@ifpackagewith{neurips_2021}{preprint}
{
\def\preprint{}
}{}
\makeatother

\usepackage[utf8]{inputenc} 
\usepackage[T1]{fontenc}    
\usepackage{url}            
\usepackage{booktabs}       
\usepackage{amsfonts}       
\usepackage{nicefrac}       
\usepackage{microtype}      
\usepackage{xcolor}         

\usepackage[colorlinks=true,citecolor=black,urlcolor=black,linkcolor=black]{hyperref}       
\usepackage{subcaption}
\usepackage{bm}
\usepackage{amsmath}
\usepackage{wrapfig}
\usepackage{placeins}

\usepackage{hyperref}       

\usepackage{tikz,pgfplots}
\usetikzlibrary{plotmarks}

\pgfplotsset{/pgf/number format/.cd, 1000 sep={}}

\pgfplotsset{every axis/.append style={
  grid style={line width=0.6pt,dotted,gray}}}

\pgfplotsset{every axis/.append style={
  legend style={inner xsep=1pt, inner ysep=0.5pt, nodes={inner sep=1pt, text depth=0.1em},draw=none,fill=none}
}}

\pgfplotsset{every axis/.append style={
  colorbar style={width=3mm,xshift=-2mm,major tick length=2pt}
}}

\newcommand{\nipstitle}[1]{{%
    \def\toptitlebar{\hrule height4pt \vskip .25in \vskip -\parskip} 
    \def\bottomtitlebar{\vskip .29in \vskip -\parskip \hrule height1pt \vskip .09in} 
    \phantomsection\hsize\textwidth\linewidth\hsize%
    \vskip 0.1in%
    \toptitlebar%
    \begin{minipage}{\textwidth}%
        \centering{\LARGE\bf #1\par}%
    \end{minipage}%
    \bottomtitlebar%
    \addcontentsline{toc}{section}{#1}%
}}

\definecolor{mycolor0}{rgb}{0.2667,0.4471,0.7098}
\definecolor{mycolor1}{rgb}{0.1647,0.6706,0.3804}
\definecolor{mycolor2}{rgb}{0.8275,0.2627,0.3059}
\definecolor{mycolor3}{rgb}{0.5216,0.4392,0.7176}
\definecolor{mycolor4}{rgb}{0.8118,0.7255,0.4118}
\definecolor{mycolor5}{rgb}{0.2745,0.7176,0.8157}

\definecolor{mylcolor0}{rgb}{0.6902,0.7686,0.8863}
\definecolor{mylcolor1}{rgb}{0.5451,0.8902,0.6941}
\definecolor{mylcolor2}{rgb}{0.9412,0.7490,0.7647}
\definecolor{mylcolor3}{rgb}{0.8627,0.8392,0.9176}
\definecolor{mylcolor4}{rgb}{0.9569,0.9373,0.8667}
\definecolor{mylcolor5}{rgb}{0.7529,0.9020,0.9373}
\definecolor{mylcolor6}{rgb}{0.8750,0.8750,0.8750}
\definecolor{mygrey}{rgb}{0.25, 0.25, 0.25}

\newcommand{\T}{\mathsf{T}}
\renewcommand{\mid}{\,|\,}

\newcommand{\mbf}[1]{\mathbf{#1}}

\newcommand{\vx}{\mbf{x}}
\newcommand{\vxHat}{\widehat{\mbf{x}}}
\newcommand{\va}{\mbf{a}}
\newcommand{\vb}{\mbf{b}}
\newcommand{\vh}{\mbf{h}}
\newcommand{\vr}{\mbf{r}}
\newcommand{\vy}{\mbf{y}}
\newcommand{\vz}{\mbf{z}}
\newcommand{\vm}{\mbf{m}}
\newcommand{\vw}{\mbf{w}}
\newcommand{\vp}{\mbf{p}}
\newcommand{\MC}{\mbf{C}}
\newcommand{\MI}{\mbf{I}}

\newcommand{\vmu}{\bm{\mu}}
\newcommand{\vtheta}{\bm{\theta}}

\newcommand{\dd}{\mathrm{d}}

\newcommand{\eg}{\textit{e.g.}}
\newcommand{\ie}{\textit{i.e.}}

\usepackage[capitalize,nameinlink]{cleveref}
\crefname{section}{Sec.}{Sects.}
\crefname{appendix}{App.}{Apps.}
\crefname{figure}{Fig.}{Figs.}
\crefname{wrapfigure}{Fig.}{Figs.}

\renewcommand{\paragraph}[1]{{\bf #1}~~}

\begin{document}


%

\title{Deep Residual Mixture Models}

\author{
  Perttu H\"am\"al\"ainen \\ Aalto University \\ Finland 
  \And 
  Martin Trapp \\ Aalto University \\ Finland 
  \And 
  Tuure Saloheimo \\ Aalto University \\ Finland 
  \And 
  Arno Solin \\ Aalto University \\ Finland 
}


\maketitle

\begin{abstract}
We propose Deep Residual Mixture Models (DRMMs), a novel deep generative model architecture. Compared to other deep models, DRMMs allow more flexible conditional sampling: The model can be trained once with all variables, and then used for sampling with arbitrary combinations of conditioning variables, Gaussian priors, and (in)equality constraints. This provides new opportunities for interactive and exploratory machine learning, where one should minimize the user waiting for retraining a model. We demonstrate DRMMs in constrained multi-limb inverse kinematics and controllable generation of animations.

\end{abstract}

\section{Introduction}
\label{sec:intro}
Deep generative models can be cumbersome for exploratory and interactive machine learning, as the conditioning variables for sampling need to be specified when training. For example, in conditional Generative Adversarial Networks (GANs, \cite{goodfellow2014}) and conditional Variational Autoencoders (VAEs, \cite{kingma2014}), the conditioning variables are implemented as additional inputs to the generator network. Changing the variables requires retraining the network. Although it is possible to train a neural model to operate on masks for the conditioning variables \cite{yu2019free}, there are only few deep generative architectures that allow post-training specification of arbitrary conditioning variables, priors, and (in)equality constraints. 

\begin{wrapfigure}{r}{0.4\textwidth}
  \centering
	\vspace{-5pt}

   \includegraphics[width=0.4\textwidth]{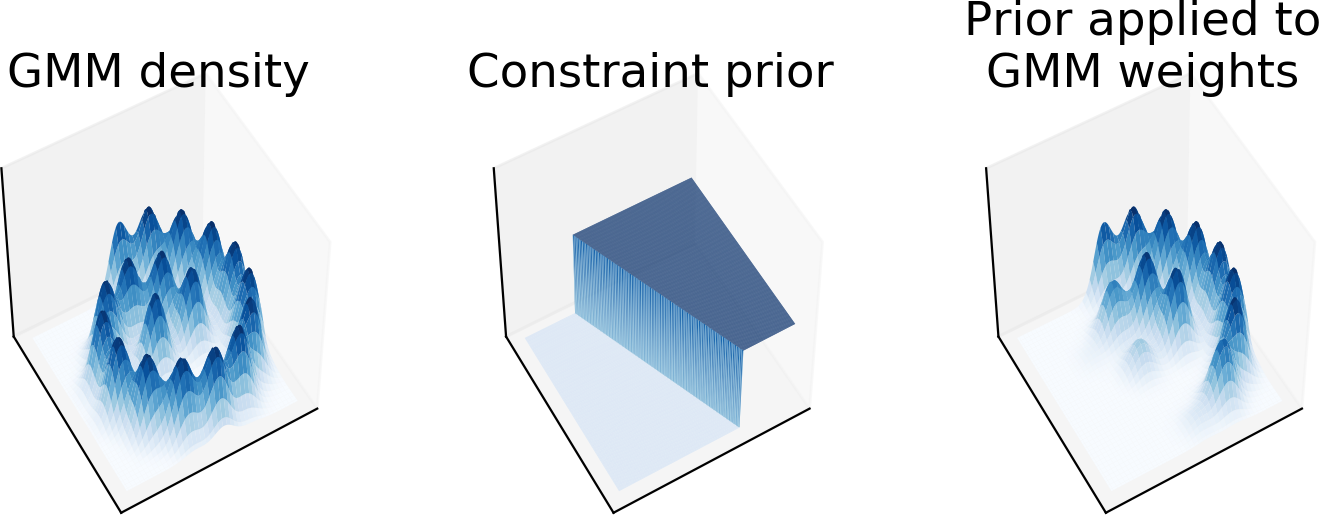}
	\vspace{-10pt}

\small
  \caption{Approximating a linear inequality constraint through GMM component weights.  } 
	\vspace{-10pt}
  \label{fig:intro_GMM}
\end{wrapfigure}

A classical machine learning tool that achieves the above is the Gaussian Mixture Model (GMM, \eg, \cite{bishop2006pattern,deisenroth2020mathematics}) which models a probability density as $p(\vx)=\sum_k w_k \mathrm{N}(\vx \mid \vmu_k, \MC_k)$, where $k$ denotes mixture component index, $w_k$ denotes component weight, and $\mathrm{N}(\vx \mid \vmu_k,\MC_k)$ denotes a Gaussian PDF with mean $\vmu_k$ and covariance $\MC_k$. Sampling from a GMM comprises two operations: Randomly select a mixture component using $w_k$ as the selection probabilities, and then draw a Gaussian sample from the selected component. As illustrated in \cref{fig:intro_GMM}, a simple way to approximate sampling priors is to use modified weights $w'_k \propto w_k c_k$, where $c_k$ is the integral of the product of a prior $q(\vx)$ and the component's Gaussian PDF, $c_k = \int q(\vx)\,\mathrm{N}(\vx \mid \vmu_k,\MC_k)\,\mathrm{d}\vx$. Sampling constraints can be implemented as priors that are 1 for valid samples and 0 for invalid samples. Simple closed-form expressions for $c_k$ exist for Gaussian priors and linear (in)equality constraints, if the GMM component covariances are isotropic, $\MC_k=\sigma^2 \MI$.

The constraint approximation above becomes more accurate as the number of mixture components  increases. However, modeling complex, high-dimensional data may require a prohibitively large number of components, and although deep and more scalable GMM variants exist (\eg, \cite{van2014factoring,viroli2019deep}), they do not allow such a simple way to incorporate priors and constraints.

\begin{figure*}[t]
  \centering
  \includegraphics[width=\textwidth]{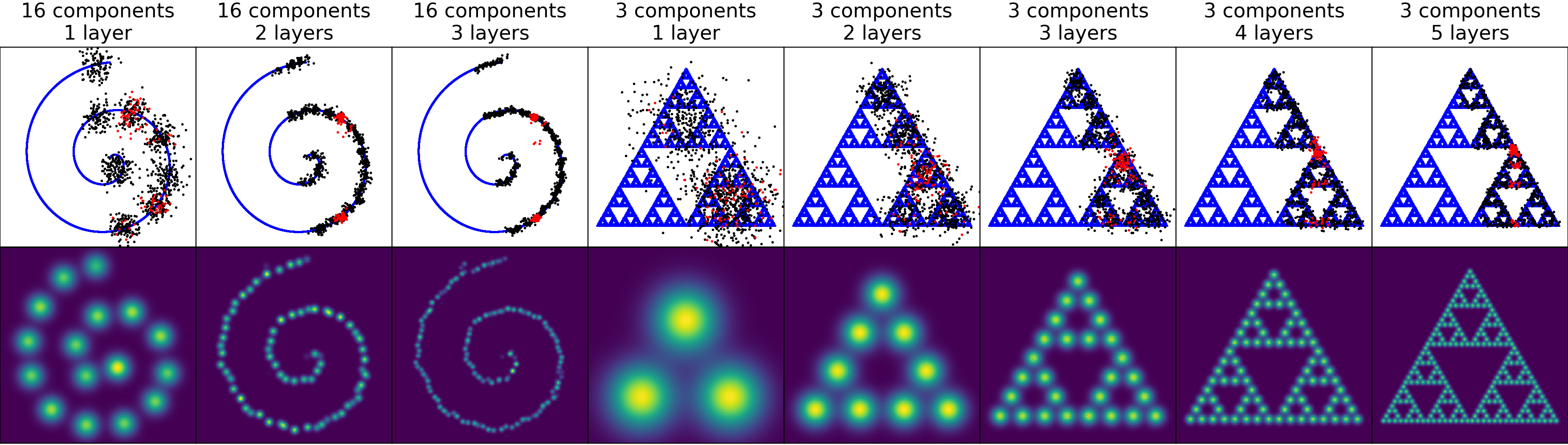}  
  \caption{DRMM samples and density (likelihood) estimates with 2D training data. Samples can be conditioned with both inequalities (black, $ax+by+c>0$) and equalities (red, $x+c=0$). DRMM capacity can grow exponentially with depth, exploiting data self-similarity: The Sierpinski triangle model has three mixture components per layer, and a $L$-layer DRMM models $3^L$ density modes. }
  \label{fig:sierpinsky}
\end{figure*}

\paragraph{Contributions}
To bridge the gap above, we propose Deep Residual Mixture Models (DRMMs), a novel deep extension of GMMs that scales to high-dimensional data while still allowing adding arbitrary priors and constraints after training. We demonstrate the approach in constrained inverse kinematics (IK) and controllable generation of animations. Examples of DRMM samples and likelihood estimates are shown in \cref{fig:sierpinsky}. The figure shows how sample quality and constraint satisfaction improve with depth. In the best case, when the data exhibits suitable self-similarity, the number of modes modeled by a DRMM can grow exponentially with depth. A TensorFlow \cite{abadi2016tensorflow} implementation of DRMM is available at 
\ifdefined\preprint
\url{https://github.com/PerttuHamalainen/DRMM}.
\else
\textit{<URL removed for review. Code and data are provided as supplementary material.>}
\fi



\section{Background} 
\label{sec:background}
We draw from the vast literature of deep, residual, and mixture models. As illustrated in \cref{fig:DRMM} and detailed in \cref{sec:methods}, a characterizing feature of our DRMM architecture is that the output of each layer is a combination of a stochastic latent variable and a modeling residual. The DRMM reduces to Residual Vector Quantization (RVQ, \cite{juang1982multiple, kossentini1995image}), if each layer only outputs the residual and the latent variable is made deterministic by assigning each input to the most probable mixture component. RVQ is a classical data compression method, predominantly known in signal processing. However, RVQ does not allow sampling, as it stores no information of which encodings are valid ones. Through the vector quantization analogue, the DRMM is also related to VQ-VAE models \cite{van2017neural} which apply vector quantization on the embeddings generated by convolutional encoder networks (see also \cite{garbacea2019low,razavi2019generating}). Many earlier approaches have utilized categorical representations with autoencoder architectures and established ways to backpropagate through samples \cite{maddison2016concrete, jang2016categorical}. The main difference to our architecture is that we do not utilize a convnet encoder, which allows us to condition the encodings with any number or combination of known variables.  

The DRMM has similar dimensionality-growing skip connections as DenseNets~\cite{huang2017densely}, but the connections are probabilistic and pass through the residual computation. DenseNets are inspired by the classical cascade-correlation networks \cite{fahlman1990cascade} where each layer concatenates its input with a new latent variable. DRMMs could be considered a probabilistic generative version of cascade correlation networks. Inequality constraints in cascade correlation networks have been considered by \citet{nobandegani2018example}, where the conditioning is implemented as a penalty term and the sampling done via MCMC, which can be computationally costly.


Extensions to plain Gaussian mixtures feature, for example, residual k-means \cite{yuan2013transform} that also increases intrinsic dimensionality, but using trees. Our model can be considered similar, with a layer per tree level, but with weight-sharing between subtrees to avoid the storage cost growing exponentially with depth. There are also hierarchical Gaussian mixture models \cite{garcia2010hierarchical}, that do not, however, feature residual connections. Each layer of a DRMM can also be considered a simple autoencoder. Traditional autoencoder stacks utilize successive per-layer encoding and decoding steps during pre-training. Yet, the decoders are discarded and a final output layer still needs to be trained, and they are also only trained on predictors in the classifier case \cite{vincent2010stacked}.

Flow-based models \cite{dinh2015nice,rezende2015variational, dinh2016density,kingma2018glow} are similar to DRMMs in that each layer progressively maps the input distribution to a simplified or`whitened' latent distribution. Latent samples can be mapped back to the input space due to the invertibility of the models. In contrast, DRMM sampling happens during the forward pass through the network and the contributions of each layer combine additively. The benefit of DRMMs over flow-based models as well as Variational Autoencoders~\cite{kingma2014} and Generative Adversarial Networks~\cite{goodfellow2014} is that a trained model can be conditioned with arbitrary variable combinations and inequality constraints, which comes at no additional cost and requires no modifications to the training. 

Probabilistic circuits, such as sum-product networks \cite{poon2011} and Einsum networks \cite{peharz2020}, are explicit likelihood models that allow arbitrary conditioning through recursive decomposition of the random variables. However, their depth is often inherently limited by the dimensionality of the data, limiting their expressiveness in certain cases. To overcome this issue, \cite{pevny2020SPTNs} introduced a combination of sum-product networks and affine flows, resulting in a deep mixture with intermediate linear change of variable operations. However, their approach requires to store and update Givens or the Householders parametrization of the unitary matrices used to represent the SVD decomposition of the transformation matrix. In contrast, DRMMs do not rely on a recursive decomposition of the random variables, nor do they need to store any transformation matrices, but they still efficiently represent an exponentially large number of mixture components.

\section{Deep Residual Mixture Models}
\label{sec:methods}

\begin{wrapfigure}{r}{0.5\textwidth}
  \centering
\vspace{-10pt}
\hspace{-10pt}
  \includegraphics[width=0.5\textwidth]{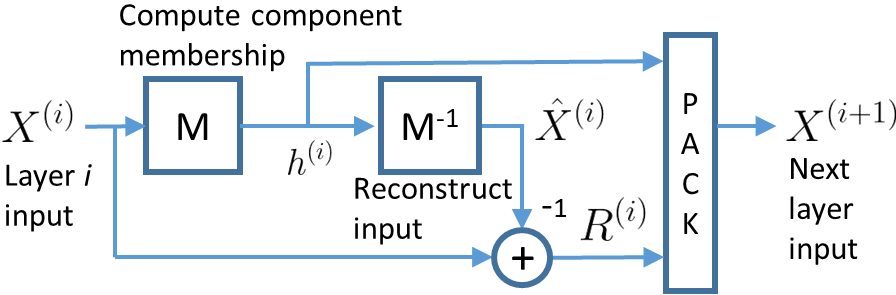}
  \caption{A DRMM layer. } 
  \label{fig:DRMM}
\vspace{-10pt}
\end{wrapfigure}

\cref{fig:DRMM} illustrates the DRMM architecture. We denote layer indices by superscripts, but drop them for brevity when not relevant.  Each DRMM layer is a mixture model that encodes the input $X$ as a categorical latent variable $h$, which denotes a mixture component membership. The M$^{-1}$ block reconstructs the input using $h$ as $\widehat{X} \approx X$, which then gives the modeling residual $R=X-\widehat{X}$.  Deep models are constructed by stacking multiple layers so that each layer observes both the residual and the latent state that produced the residual of the previous layer. 


\paragraph{Data Stream Notation}
To allow layers to observe and process the residuals, latents, and their subsequent residuals in a unified manner, we utilize the concept of {\em data streams}: Each input data point is a tuple  $X=[\vx_1,\vx_2,\ldots]$, where the subscripts are stream indices and the vectors $\vx$ represent either real-valued multivariate data or log-probability distributions over a categorical variable. Similarly, $\widehat{X}=[\widehat{\vx}_1,\widehat{\vx}_2,\ldots]$, and $R=[\vr_1,\vr_2,\ldots]=X-\widehat{X}=[\vx_1-\widehat{\vx}_1,\vx_2-\widehat{\vx}_2,\ldots]$. The latent state $h$ of each layer becomes an additional categorical input stream for the next layer as $X^{(i+1)}=[\vx_1^{(i+1)},\vx_2^{(i+1)},\ldots]=[\vr_1^{(i)},\vr_2^{(i)},\ldots,\vh^{(i)}]$, where $\vh^{(i)}=\log (\mathrm{smooth}(\mathrm{onehot}(h^{(i)})))$ and $\mathrm{smooth}(\cdot)$ denotes Laplace smoothing.

\begin{figure*}[t]
\centering
\includegraphics[width=\textwidth]{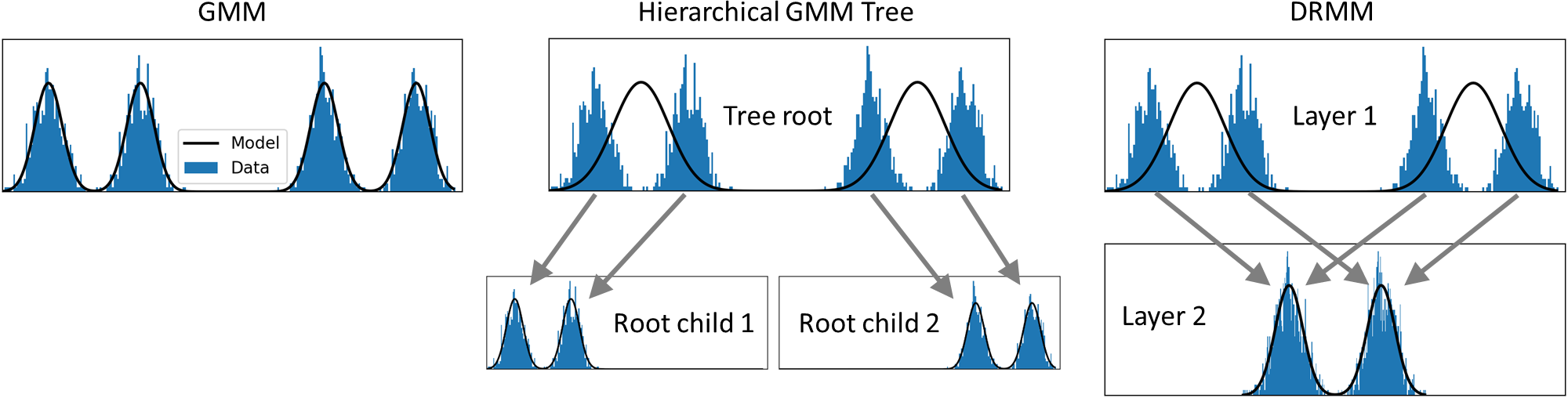}
\caption{A DRMM layer assigns each input vector to a mixture component, similar to a node of a tree-based hierarchical GMM. However, instead of child nodes receiving subsets of the data, all the data is transformed to the residual space and passed to a single next layer.}
\label{fig:drmm_vs_gmm}
\end{figure*}

\paragraph{Intuition} DRMM layers progressively extract more information from the input and encode the information into the latent variables. This allows deeper models to build richer latent representations. As illustrated in \cref{fig:drmm_vs_gmm}, a DRMM layer is conceptually similar to a node of a tree-based hierarchical GMM. However, instead of having multiple children per node, we have  only a single next layer. The next layer can construct a more precise model with the same number of components, provided that multiple input density modes map to a single residual density mode. 

\subsection{Single Layer Likelihood}
DRMMs are trained in Maximum Likelihood fashion, by maximizing the average $\log p(X^{(1)} \mid \vtheta)$, where $\vtheta$ denotes model parameters. We first detail the likelihood model of a single layer, which we later extend to deep models in \cref{sec:deeplikelihood}. 

Each DRMM layer defines a mixture distribution of $K$ components over parametric families $f_s$, factorized over all input streams $s=1,2,\ldots,S$: 
\begin{equation}
  p(X \mid \vtheta) = \sum_{h=1}^{K} \mathrm{Pr}(h \mid \vtheta) \, p(X \mid h,\vtheta)=\sum_{h=1}^K w_h \prod_{s=1}^{S} f_{s}(\vx_s \mid h,\vtheta_s), \label{eq:px} 
\end{equation}
where $\vtheta =[w_1,\ldots,w_K,\vtheta_1,\ldots,\vtheta_S]$ and $\vtheta_s$ denotes parameters specific to stream $s$. We assume that component weights $\vw^{(L)} \in \Delta^{K^{(L)}-1}$. Depending on the data-type of the stream $s$, $f_{s}$ either takes the form of an isotropic Gaussian density (real-valued streams), \ie,
\begin{equation}
    f_{s}(\vx_s \mid h,\vtheta_s) =\mathrm{N}(\vx_s \mid \vmu_{h,s},\sigma_s^2\, \MI) \, , \label{eq:gaussianpx}
\end{equation}
or of a categorical distribution (discrete streams, \eg, latent encodings), \ie,
\begin{equation}
    f_{s}(\vx_s \mid h,\vtheta_s) = \prod_{i} p_{h,s,i}^{y_i} \, , \label{eq:categorical}
\end{equation}
where $\vy = \mathrm{softmax}(\vx)$ and $\vp_{h,s}$ is a vector of probabilities. 

To avoid computational precision issues, we operate in log-domain, where the product in \cref{eq:px} becomes a summation that can be divided into two parts:
\begin{multline}
\log \prod_{s=1}^{S} f_{s}(\vx_s \mid h,\vtheta_s) = \sum_{r \in \mathcal{S}_\mathrm{real}}  \left( -\frac{1}{2 \sigma_r^2} \|\vx_r-\vmu_{h,r}\|^2  -  N_r \log \sigma_r - \frac{N_r}{2}\log 2\pi \right) \\
 + \sum_{c \in \mathcal{S}_\mathrm{cat}} \mathrm{softmax}(\vx_c)^\T \log \vp_{h,c}, \label{eq:logmembership}
\end{multline}
where $\mathcal{S}_\mathrm{real}$ and $\mathcal{S}_\mathrm{cat}$ are the sets of real-valued and categorical input streams and $N_r$ is the number of elements in $\vx_r$. 

It should be noted that the categorical distribution in \cref{eq:categorical} is correctly normalized only if $\vy=\mathrm{softmax}(\vx)$ is one-hot. However, if one interprets $\vy$ as describing a population of categorical samples, the cross-entropy term in \cref{eq:logmembership} gives the correct average log-likelihood for the population. Accordingly, the residual describes the population of reconstruction errors. The population interpretation is also compatible with the real-valued $\vx$, if treating $\vx$ as the population mean and assuming zero population variance.  


\subsection{The M and M$^{-1}$ Blocks}\label{sec:memberships}
The M-block in \cref{fig:DRMM} samples $h \sim \mathrm{Pr}(h \mid X,\vtheta) \propto Pr(h \mid \vtheta) \, p(X \mid h,\vtheta)=w_h \prod_{s=1}^{S} f_{s}(\vx_s \mid h,\vtheta_s)$. The M$^{-1}$ block maps the $h$ back to input domain $\widehat{X}=[\widehat{\vx}_1,\widehat{\vx}_2,\ldots]$, which then gives the residual as $R=X-\widehat{X}$. For a real-valued input stream, $\widehat{\vx}_i=\vmu_{h,i}$. For categorical streams,  $\widehat{\vx}_i=\vp_{h,i}$. 

\subsection{Likelihood for a Deep Model}\label{sec:deeplikelihood}
A multilayer model's likelihood for the observable inputs $X^{(1)}$ requires a marginalization over all the latents:

\begin{equation}
p(X^{(1)} \mid \vtheta)=\sum_{h^{(1)}} \ldots \sum_{h^{(L)}}  p(X^{(1)},h^{(1)},\ldots,h^{(L)} \mid \vtheta). \label{eq:genericlikelihood}
\end{equation}
  
This is not tractable for arbitrary $L$, but as shown below, one can approximate the likelihood through sampling. In particular, we can sample a latent assignment $h$ for each layer in a single forward pass and then evaluate the joint, which {\em reduces to evaluating the likelihood of the last layer}. 

First, recall that a discrete/categorical random variable can equivalently be considered a continuous random variable with a Dirac mixture density. The `pack' operation explicitly maps $h$ to such a Dirac mixture in the $\vh$ space. Therefore, we may equate:  

\begin{equation}
p(X^{(i+1)})=p([R^{(i)},\vh^{(i)}])=p(R^{(i)},h^{(i)}).
\end{equation}

The residual computation $R=X-\widehat{X}$ is a simple shift operation that does not squeeze or expand the distribution of $X$ and the shift $\widehat{X}$ is constant given $h$. In other words, the mapping $R=f_h(X)=X-\widehat{X}$ is invertible with $| \det \nabla_R f_h^{-1}(R)| =1$ and we have:

\begin{equation}
p(R^{(i)},h^{(i)})=p(X^{(i)},h^{(i)}). \label{eq:pnext}
\end{equation}

Hence, $p(X^{(i+1)})=p(X^{(i)},h^{(i)})$, which can be used to recursively obtain:

\begin{align}
p(X^{(1)},h^{(1)},\ldots,h^{(L)} \mid \vtheta)=p(X^{(2)},h^{(2)},\ldots,h^{(L)} \mid \vtheta)=\ldots=p(X^{(L)},h^{(L)} \mid \vtheta).
\end{align}

Substituting this to \cref{eq:genericlikelihood} and writing the summation as an expectation over $h^{(l)}\sim q(h^{(l)})$, we obtain: 

\begin{align}
p(X^{(1)} \mid \vtheta)&=\sum_{h^{(1)}} \ldots \sum_{h^{(L-1)}} p(X^{(L)} \mid \vtheta) \label{eq:likelihood2}\\
&=\mathbb{E}_{h^{(1)} \sim q(h^{(1)}),\ldots,h^{(L)} \sim q(h^{(L-1)})} \left[\frac{1}{q(h^{(1)})\cdots q(h^{(L-1)})} p(X^{(L)} \mid \vtheta) \right]. \label{eq:likelihood}
\end{align}

In practice, the expectation is approximated as an average over samples. \cref{fig:sierpinsky} demonstrates this using 32 samples. As explained in \cref{sec:memberships}, we use $q(h^{(l)}) \propto w_{h^{(l)}} \prod_{s=1}^{S} f_{s}(\vx_s^{(l)} \mid h^{(l)},\vtheta_s^{(l)})$. 

The likelihood of \cref{eq:likelihood2} equals the last layer's $K$-component mixture evaluated for all the $K^{L-1}$ possible $X^{(L)}$ produced by the latents of the previous layers. This results in the up to $K^L$ likelihood modes visible in \cref{fig:sierpinsky}.

\subsection{Training}
\label{sec:training}
\label{sec:regularization}
To train DRMMs, we maximize the average log-likelihood $\frac{1}{N} \sum_{i=1}^{N} \log p(X_i^{(1)} | \vtheta)$, where the sum is over training samples. We implement this using randomly sampled minibatches of data, single-sample estimates of \cref{eq:likelihood}, and taking an Adam \cite{kingma2014adam} optimization step per minibatch. 

Directly optimizing the likelihood is prone to getting stuck at local optima. We mitigate this using curriculum learning \cite{bengio2009curriculum} and a 3-stage curriculum: 
\begin{description}
\item[Stage 1:] Pretrain to maximize a proxy objective $\sum_{l=1}^L \mathcal{L}_l$, where $\mathcal{L}_l$ denotes the log-likelihood of a DRMM constituted by the first $l$ layers. The rationale for this is that shallower models are easier to to optimize, and training the first layers as a shallow model provides a good initialization for a deeper model. We also stop gradients between layers and simplify the likelihood by omitting the $w_h$ and the categorical latent stream terms.  
\item[Stage 2:] Continue pretraining, but including the categorical latent streams in the likelihood. This makes the optimization objective more complex, but is required to allow modeling the probability of different combinations of $h$. 
\item[Stage 3:] Remove the gradient stops, include the $w_h$, and linearly interpolate from the proxy objective to the true log-likelihood. We also drop Adam learning rate by a factor of 0.1 and then linearly anneal it to zero during the stage. 
\end{description}

\begin{figure*}[t]
  \centering
  \includegraphics[width=\textwidth]{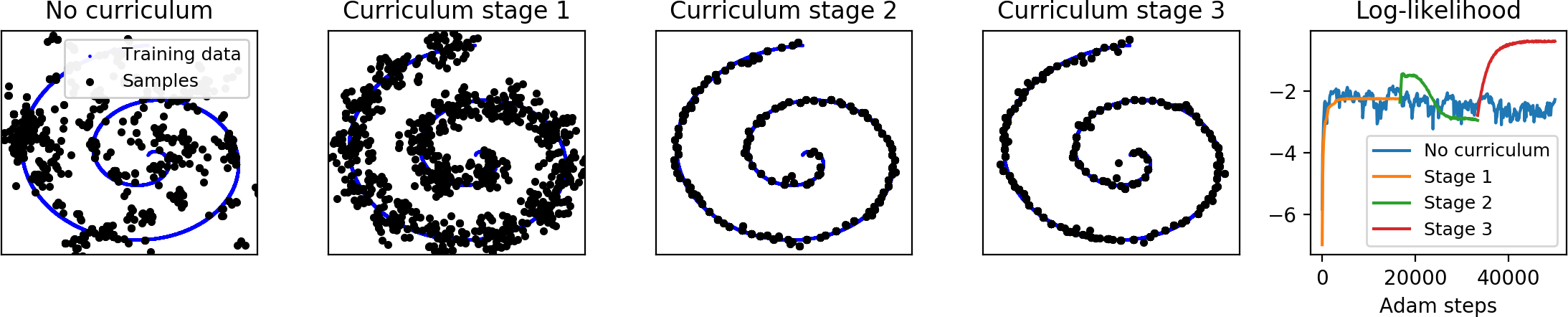}
  \caption{Effect of the training curriculum on sample quality and log-likelihood. Samples are visualized at the end of each training stage. The model has 4 layers, 16 components per layer.}
  \label{fig:curriculum}
\end{figure*}

The effect of the curriculum is visualized in \cref{fig:curriculum}. Each stage uses one third of a given total training iteration budget. Stage 1 ensures that samples and mixture components cover the training data, stage 2 increases sample precision, and stage 3 further increases the log-likelihood, although there is no clear difference in visual sample quality in this simple 2D example. 

To prevent `orphan' components that get assigned no data, stages 1 and 2 also use a regularization loss term that penalizes the distance of each component to its closest input: $\frac{1}{K} \sum_{h=1}^{K} \alpha \max_i \log p(X_i \mid h,\vtheta)$, with $\alpha = 0.001$. It is also noteworthy that the residual connections allow gradient propagation between layers, despite the sampled $h$. We did also test propagating gradients directly through the samples using reparameterized Gumbel-Softmax \cite{maddison2016concrete,jang2016categorical}, but found this to be unstable in our case. 

\subsection{Sampling}\label{sec:conditioning}
To sample from a trained model, one feeds in data $X$ and samples the $h \sim Pr(h \mid X,\vtheta)$ for all $L$ layers. A corresponding input-space sample is obtained as $\sum_{l=1}^{L} \widehat{X}^{(l)}$. Optionally, one may add Gaussian noise using the last layer's $\sigma^2$ to approximate the variance not modeled by the $h$. The noise is added in \cref{fig:sierpinsky} to make the samples consistent with the likelihood plots, but we omit it elsewhere. In effect, each sampled $\sum_{l=1}^{L} \widehat{X}^{(l)}$ is an input-space \textit{multilayer mixture component mean}. 

\paragraph{Known/conditioning variables} Because of the isotropic covariances and the factorization over streams, the contributions of each stream and variable combine additively in \cref{eq:logmembership}. Thus, to allow conditioning $h\sim w_h \prod_{s=1}^{S}f_{s}(\vx_s \mid h,\vtheta_s)$ on only a subset of variables, we can augment \cref{eq:logmembership} with per-stream and per-variable multipliers $m_{r,n}$ and $m_c$: 
\begin{multline}
\log \prod_{s=1}^{S} f_{s}(\vx_s \mid h,\vtheta_s) =  \sum_{r \in \mathcal{S}_\mathrm{real}} \sum_{n=1}^{N_r} m_{r,n} \left( -\frac{1}{2 \sigma_r^2} (x_{r,n}-\mu_{h,r,n})^2 - \log \sigma_r - \log 2\pi \right) \\
+  \sum_{c \in \mathcal{S}_\mathrm{cat}} m_{c}\: \mathrm{softmax}(\vx_c)^\T \log \vp_{h,c}. \label{eq:multipliers}
\end{multline}
Unknown/sampled variables have $m=0$ and no effect on $h$. Known/conditioning variables have $m=1$ as a default, but other values can be used to adjust the confidence/weight of each variable. The multipliers of an input stream also apply to the corresponding residual output stream.

In case all multipliers are either zero or one, \cref{eq:multipliers} can be understood as marginalising out variables for which $m=0$ and conditioning on the remaining variables for which $m=1$. See \cref{app:marginalisation} for details.

\paragraph{Priors and Constraints} Additional priors and constraints can be added similar to a conventional GMM. As discussed in \cref{sec:intro}, for a mixture component with mean $\vmu$ and covariance $\sigma^2 \MI$, a prior $q(\vx)$ is implemented as a component weight multiplier $c = \int q(\vx)\,\mathrm{N}(\vx \mid \vmu,\sigma^2 \MI)\,\mathrm{d}\vx$. A constraint is considered a prior that is zero where the constraint is not satisfied. 

The priors are applied layerwise and transformed to the residual output space for the next layer. In effect, {\em the priors flow through the network with the data} and inform each layer's $h$. Substituting $\vx=\vr+\vxHat$ to the constraint equality $\va^\top\vx+b>0$, one gets the residual constraint $\va_r^\top\vr+b_r > 0$, where $\va_r=\va, b_r=b+\va^\top\vxHat$. The same can be applied to equalities $\va^\top\vx+b=0$. For Gaussian priors, one only needs to shift the prior mean by $-\vxHat$, similar to each input vector.  

Because of the spherical/isotropic component covariances, a linear inequality constraint only needs to be integrated in 1D, along the constraint hyperplane normal. We project the component mean $\vmu$ on the normal as $v=\va^\top\vmu+b$, assuming normalized constraint representation with $||\va||=1$. We integrate the component Gaussian on the valid side of the hyperplane, along the normal, given by the Gaussian CDF formula $0.5 (1 + \mathrm{erf}(v/(\sigma \sqrt{2})))$. Box constraints are implemented as per-variable linear inequalities $x_i+b_i>0$. For Gaussian priors, we can use the product of $\int  \mathrm{N}(x \mid \mu_1,\sigma_1^2)\mathrm{N}(x \mid \mu_2,\sigma_2^2) \,\mathrm{d}x =  \mathrm{N}(\mu_1 \mid \mu_2,\sigma_1^2 + \sigma_2^2)$ for each variable, substituting mixture component mean and standard deviation for $\mu_1,\sigma_1$ and prior mean and standard deviation for $\mu_2,\sigma_2$. This assumes diagonal covariance for the prior. 

\subsection{Scaling with Depth}\label{sec:depth}
With $L$ layers and $K$ components per layer, there are $K^L$ possible latent variable combinations, each corresponding to a multilayer mixture component mean $\sum_{l=1}^{L} \widehat{X}^{(l)}$ (see \cref{sec:conditioning}). However, unless the data is suitably self-similar like the Sierpinski fractal in \cref{fig:sierpinsky}, some of the components may lie outside the data manifold, and will be assigned low sampling probabilities through the weights $w_h$ and the categorical input stream terms in \cref{eq:logmembership} and \cref{eq:multipliers}. Preventing this, as done in the curriculum stage 1, allows the off-manifold samples shown in \cref{fig:curriculum}. In other words, how many of the $K^L$ components are useful in practice depends on how well the data matches DRMM's inductive bias of self-similarity.  

The number of model parameters grows quadratically with depth. Denoting the sum total of input stream variables as $D$, a DRMM component has $D$ parameters for the $\vmu,\vp$ and one parameter for $w$. A layer has $K$ components and a scalar $\sigma$, assuming a single real-valued 1\textsuperscript{st} layer input stream. Thus, a layer has $(D+1)K+1 \approx DK$ parameters. With $\vh \in \mathbb{R}^K$ appended to layer outputs, dimensionality grows as $D_l=D_1+(l-1)K$, where $l$ is layer index. Substituting this to $DK$ and summing over $L$ layers results in a total of $D_1 L K+0.5 L(L-1)K^2$ model parameters, or $\mathcal{O}(L^2)$.

\section{Experiments}\label{sec:experiments}
The experiments below demonstrate the uniquely flexible sampling of DRMMs in two applications. We also quantitatively verify that the benefits of depth outweigh the growth in model size. 

\subsection{Constrained Inverse Kinematics}\label{sec:IK}
\begin{figure*}[t]
  \centering
  \includegraphics[width=\columnwidth]{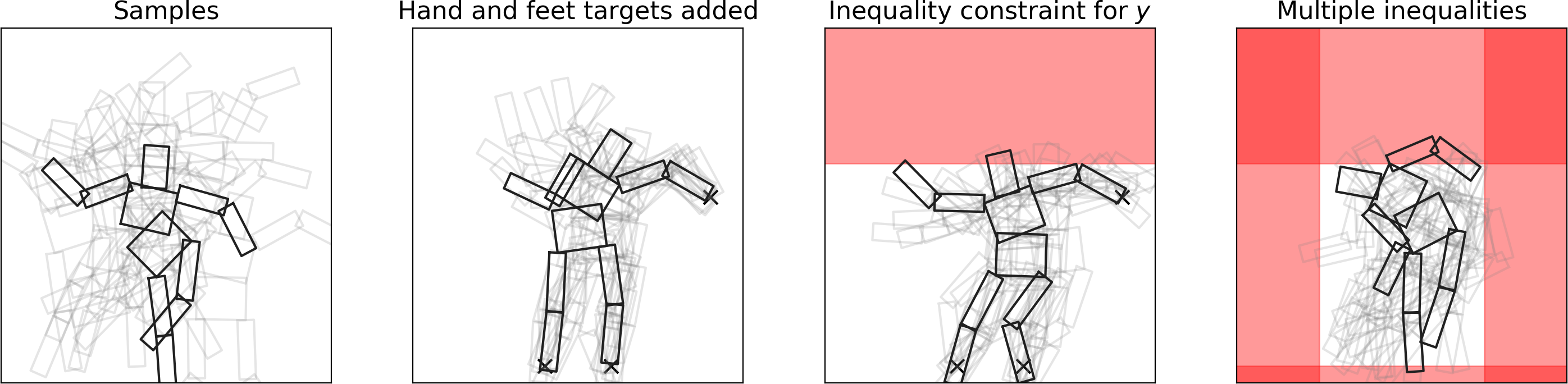}
  \caption{10-layer DRMM samples of humanoid root position, rotation, and joint angles, conditioned on goals and inequality constraints for head, hands, and/or feet. Uniquely, our model allows adding any combination of goals and constraints after training. \label{fig:ik}}

  \centering
  \vspace{14pt}
  \includegraphics[width=\columnwidth]{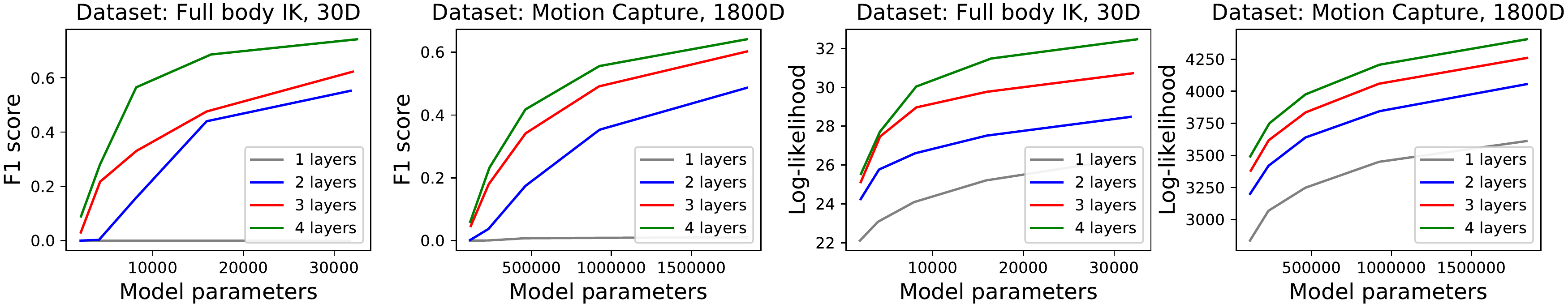}
  \caption{Sample quality (F1 score) and training data log-likelihood w.r.t. DRMM depth and the number of model parameters. Deeper models perform consistently better. \label{fig:benchmark}}
\end{figure*}

\cref{fig:ik} demonstrates DRMM forward sampling in inverse kinematics, a common problem in robotics and computer animation. IK solvers output the skeletal joint angles given one or more end-effector goal positions and constraints. This is only simple for chains of two bones, and iterative optimization and/or machine learning methods are needed for more complex skeletal structures \cite{grochow2004style,aristidou2009inverse,koker2013genetic}. Typically, multiple solutions exist for each IK problem, and it would be beneficial to have a generative model that can sample alternative solutions. 

We train a DRMM with 10 layers and 256 components per layer, using 1M random skeletal configurations of the 2D humanoid in \cref{fig:ik}. The training data are 30D vectors comprising root position and rotation, joint angles, and the world coordinates of hands, feet, top of the head, and center of mass. As illustrated in \cref{fig:ik}, the model correctly infers both root and joint parameters to approximately satisfy the goals and constraints. The samples are not perfect, but they are good enough to allow correction using an established local IK solver such as CCD \cite{wang1991combined,kenwright2012inverse}. \cref{fig:ik} visualizes the 10 samples with the highest likelihood from a batch of 64, the most likely sample in black. We know of no other machine learning based IK solution that allows one to add an arbitrary number of goals and constraints after only training once with random data. 
 
\subsection{Animation Synthesis}\label{sec:animation}
The supplemental material provides video examples of animation synthesis where a 4-layer DRMM with 512 components per layer is trained with short movement sequences extracted from the Ubisoft LaForge motion capture dataset \cite{harvey2020robust} (license: CC BY-NC-ND 4.0). Animation frames are encoded as vectors of character joint coordinates. The model is used to autoregressively sample the next poses, conditioned on both previous poses and movement goals. The same model can easily be conditioned with combinations of goals. The videos demonstrate this in two cases: 1) specifying both desired position (a flag that appears at random location) and desired facing direction (facing towards the goal), and 2) only specifying the desired position. As one would expect, the latter case makes more varied movements emerge, \eg, walking backwards and hopping sideways towards the goal. Implementation and training data details are given in \cref{app:details}. 

\begin{wrapfigure}{r}{0.4\textwidth}
  \centering
	\vspace{-5pt}
   \includegraphics[width=0.4\textwidth]{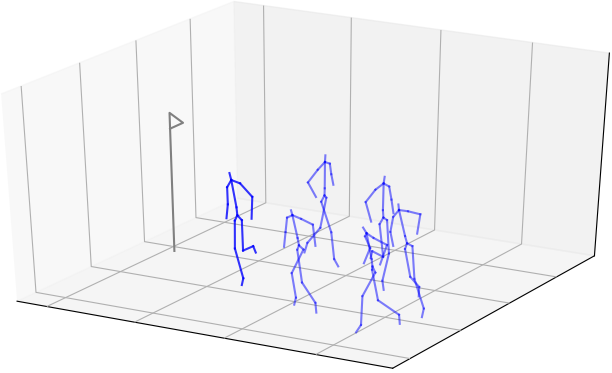}
  \label{fig:animation}
\end{wrapfigure}
Qualitatively, the results are good considering that we do not employ techniques like explicit annotation and processing of foot contacts to prevent unnaturally sliding feet \cite{harvey2020robust}. Our training clips also contain some intended sliding at rapid turns, which is faithfully reproduced by the model. The generated poses are also natural, without any explicit regularization such as implementing the forward kinematics as part of the compute graph \cite{pavllo2019modeling}. In addition to generating animations as such, our model should be directly applicable in Deep Reinforcement Learning (DRL) for motion control, as a more flexible substitute for the kinematic reference trajectory generators utilized by state-of-the-art architectures \cite{bergamin2019drecon,won2020scalable}.

\subsection{Validating the Benefits of Depth}\label{sec:gmmcompare}
DRMM's quadratic parameter growth with depth, combined with the need for self-similar data for best-case deep model performance pose a research question: {\em With real-world data, do deep models yield better performance with respect to model size?} To investigate this, \cref{fig:benchmark} compares DRMMs of various depths and sizes. Deeper models do perform consistently better, at least with our datasets. For the 1-layer model, we train from 64 to 1024 mixture components. For the deeper models, we select component counts that yield similar model sizes. The IK data is the same as in \cref{sec:IK}. The motion capture data is all the 150k 1-second segments (30 frames, 60 variables per frame) in the dataset of \cref{sec:animation}. Sample quality is measured as F1 scores---harmonic mean of precision and recall---using the approach of \citet{kynkaanniemi2019improved}.

\section{Limitations}\label{sec:limitations}
DRMM performance depends greatly on how well the residuals of data assigned to each mixture component align in a way that allows the next layer to model a more simple distribution. Presently, the residual computation is a very simple geometric transform, and we hypothesize that there are other suitable transforms that improve the alignment. For instance, one might learn a per-component scaling and rotation of the residuals that maximizes the alignment. On the other hand, this makes handling priors and constraints more complicated, as they would need to be similarly scaled an rotated. For example, a simple Gaussian prior with diagonal covariance would need a full covariance matrix after a rotation.
\section{Conclusion}
We have presented the DRMM, a generative model that combines the benefits of classical GMMs with the (best-case) exponential capacity growth of modern deep architectures. We contribute through the novel residual connection and latent variable augmentation architecture, which allows extremely flexible sample conditioning and constraining without re-training the model. In essence, one can train a DRMM without {\it a~priori} knowledge of multivariate relations, and then {\em predict anything from anything}. We demonstrate this in constrained inverse kinematics and controllable animation synthesis. It should also be possible to combine DRMMs with neural architectures that utilize mixture models or vector quantization as a building block \cite{ge2015modelling,van2017neural}, substituting a DRMM as a more scalable and versatile alternative.

\ifdefined\preprint
\begin{ack}
This work has been supported by Academy of Finland grant 299358, Technology Industries of Finland Centennial Foundation, and the computing resources provided by Aalto University's Triton computing cluster. 

\end{ack}
\fi

\begingroup
\small
\bibliographystyle{unsrtnat}
\bibliography{bibliography}

\begin{thebibliography}{41}
\providecommand{\natexlab}[1]{#1}
\providecommand{\url}[1]{\texttt{#1}}
\expandafter\ifx\csname urlstyle\endcsname\relax
  \providecommand{\doi}[1]{doi: #1}\else
  \providecommand{\doi}{doi: \begingroup \urlstyle{rm}\Url}\fi

\bibitem[Goodfellow et~al.(2014)Goodfellow, Pouget-Abadie, Mirza, Xu,
  Warde-Farley, Ozair, Courville, and Bengio]{goodfellow2014}
Ian Goodfellow, Jean Pouget-Abadie, Mehdi Mirza, Bing Xu, David Warde-Farley,
  Sherjil Ozair, Aaron Courville, and Yoshua Bengio.
\newblock Generative adversarial nets.
\newblock In \emph{Advances in Neural Information Processing Systems 27
  (NIPS)}, pages 2672--2680. Curran Associates, Inc., 2014.

\bibitem[Kingma and Welling(2014)]{kingma2014}
Diederik~P. Kingma and Max Welling.
\newblock Auto-encoding variational {B}ayes.
\newblock In \emph{International Conference on Learning Representations
  (ICLR)}, 2014.

\bibitem[Yu et~al.(2019)Yu, Lin, Yang, Shen, Lu, and Huang]{yu2019free}
Jiahui Yu, Zhe Lin, Jimei Yang, Xiaohui Shen, Xin Lu, and Thomas Huang.
\newblock Free-form image inpainting with gated convolution.
\newblock In \emph{International Conference on Computer Vision (ICCV)}, pages
  4470--4479. IEEE/CVF, 2019.

\bibitem[Bishop(2006)]{bishop2006pattern}
Christopher~M Bishop.
\newblock \emph{Pattern Recognition and Machine Learning}.
\newblock Springer, 2006.

\bibitem[Deisenroth et~al.(2020)Deisenroth, Faisal, and
  Ong]{deisenroth2020mathematics}
Marc~Peter Deisenroth, A~Aldo Faisal, and Cheng~Soon Ong.
\newblock \emph{Mathematics for Machine Learning}.
\newblock Cambridge University Press, 2020.

\bibitem[van~den Oord and Schrauwen(2014)]{van2014factoring}
Aaron van~den Oord and Benjamin Schrauwen.
\newblock Factoring variations in natural images with deep {G}aussian mixture
  models.
\newblock In \emph{Advances in Neural Information Processing Systems 27
  (NIPS)}, pages 3518--3526. Curran Associates, Inc., 2014.

\bibitem[Viroli and McLachlan(2019)]{viroli2019deep}
Cinzia Viroli and Geoffrey~J McLachlan.
\newblock Deep {G}aussian mixture models.
\newblock \emph{Statistics and Computing}, 29\penalty0 (1):\penalty0 43--51,
  2019.

\bibitem[Abadi et~al.(2016)Abadi, Barham, Chen, Chen, Davis, Dean, Devin,
  Ghemawat, Irving, Isard, et~al.]{abadi2016tensorflow}
Mart{\'\i}n Abadi, Paul Barham, Jianmin Chen, Zhifeng Chen, Andy Davis, Jeffrey
  Dean, Matthieu Devin, Sanjay Ghemawat, Geoffrey Irving, Michael Isard, et~al.
\newblock Tensorflow: a system for large-scale machine learning.
\newblock In \emph{OSDI}, volume~16, pages 265--283, 2016.

\bibitem[Juang and Gray(1982)]{juang1982multiple}
Biing-Hwang Juang and A~Gray.
\newblock Multiple stage vector quantization for speech coding.
\newblock In \emph{IEEE International Conference on Acoustics, Speech, and
  Signal Processing (ICASSP)}, volume~7, pages 597--600. IEEE, 1982.

\bibitem[Kossentini et~al.(1995)Kossentini, Smith, and
  Barnes]{kossentini1995image}
Faouzi Kossentini, Mark~JT Smith, and Christopher~F Barnes.
\newblock Image coding using entropy-constrained residual vector quantization.
\newblock \emph{IEEE Transactions on Image Processing}, 4\penalty0
  (10):\penalty0 1349--1357, 1995.

\bibitem[van~den Oord et~al.(2017)van~den Oord, Vinyals, and
  Kavukcuoglu]{van2017neural}
Aaron van~den Oord, Oriol Vinyals, and Koray Kavukcuoglu.
\newblock Neural discrete representation learning.
\newblock In \emph{Advances in Neural Information Processing Systems 30
  (NIPS)}, pages 6306--6315. Curran Associates, Inc., 2017.

\bibitem[G{\^a}rbacea et~al.(2019)G{\^a}rbacea, van~den Oord, Li, Lim, Luebs,
  Vinyals, and Walters]{garbacea2019low}
Cristina G{\^a}rbacea, A{\"a}ron van~den Oord, Yazhe Li, Felicia~SC Lim,
  Alejandro Luebs, Oriol Vinyals, and Thomas~C Walters.
\newblock Low bit-rate speech coding with {VQ-VAE} and a {WaveNet} decoder.
\newblock In \emph{IEEE International Conference on Acoustics, Speech, and
  Signal Processing (ICASSP)}, pages 735--739. IEEE, 2019.

\bibitem[Razavi et~al.(2019)Razavi, van~den Oord, and
  Vinyals]{razavi2019generating}
Ali Razavi, Aaron van~den Oord, and Oriol Vinyals.
\newblock Generating diverse high-fidelity images with {VQ-VAE-2}.
\newblock In \emph{Advances in Neural Information Processing Systems 32
  (NeurIPS)}, pages 14837--14847. Curran Associates, Inc., 2019.

\bibitem[Maddison et~al.(2017)Maddison, Mnih, and Teh]{maddison2016concrete}
Chris~J Maddison, Andriy Mnih, and Yee~Whye Teh.
\newblock The concrete distribution: {A} continuous relaxation of discrete
  random variables.
\newblock In \emph{International Conference on Learning Representations
  (ICLR)}, 2017.

\bibitem[Jang et~al.(2017)Jang, Gu, and Poole]{jang2016categorical}
Eric Jang, Shixiang Gu, and Ben Poole.
\newblock Categorical reparameterization with {G}umbel-{S}oftmax.
\newblock In \emph{International Conference on Learning Representations
  (ICLR)}, 2017.

\bibitem[Huang et~al.(2017)Huang, Liu, Van Der~Maaten, and
  Weinberger]{huang2017densely}
Gao Huang, Zhuang Liu, Laurens Van Der~Maaten, and Kilian~Q Weinberger.
\newblock Densely connected convolutional networks.
\newblock In \emph{Proceedings of the IEEE Conference on Computer Vision and
  Pattern Recognition (CVPR)}, pages 4700--4708, 2017.

\bibitem[Fahlman and Lebiere(1990)]{fahlman1990cascade}
Scott~E Fahlman and Christian Lebiere.
\newblock The cascade-correlation learning architecture.
\newblock In \emph{Advances in Neural Information Processing Systems 2 (NIPS)},
  pages 524--532. Morgan-Kaufmann, 1990.

\bibitem[Nobandegani and Shultz(2018)]{nobandegani2018example}
Ardavan~Salehi Nobandegani and Thomas~R Shultz.
\newblock Example generation under constraints using cascade correlation neural
  nets.
\newblock In \emph{40th Annual Cognitive Science Society Meeting (CogSci)},
  pages 2388--2393, 2018.

\bibitem[Yuan and Liu(2013)]{yuan2013transform}
Jiangbo Yuan and Xiuwen Liu.
\newblock Transform residual k-means trees for scalable clustering.
\newblock In \emph{IEEE 13th International Conference on Data Mining
  Workshops}, pages 489--496. IEEE, 2013.

\bibitem[Garcia et~al.(2010)Garcia, Nielsen, and Nock]{garcia2010hierarchical}
Vincent Garcia, Frank Nielsen, and Richard Nock.
\newblock Hierarchical {G}aussian mixture model.
\newblock In \emph{IEEE International Conference on Acoustics, Speech, and
  Signal Processing (ICASSP)}, pages 4070--4073, 2010.

\bibitem[Vincent et~al.(2010)Vincent, Larochelle, Lajoie, Bengio, and
  Manzagol]{vincent2010stacked}
Pascal Vincent, Hugo Larochelle, Isabelle Lajoie, Yoshua Bengio, and
  Pierre-Antoine Manzagol.
\newblock Stacked denoising autoencoders: {L}earning useful representations in
  a deep network with a local denoising criterion.
\newblock \emph{Journal of Machine Learning Research}, 11\penalty0
  (110):\penalty0 3371--3408, 2010.

\bibitem[Dinh et~al.(2015)Dinh, Sohl-Dickstein, and Bengio]{dinh2015nice}
Laurent Dinh, Jascha Sohl-Dickstein, and Samy Bengio.
\newblock {NICE}: {N}on-linear independent components estimation.
\newblock In \emph{International Conference on Learning Representations
  (ICLR)}, 2015.

\bibitem[Rezende and Mohamed(2015)]{rezende2015variational}
Danilo Rezende and Shakir Mohamed.
\newblock Variational inference with normalizing flows.
\newblock In \emph{Proceedings of the 32nd International Conference on Machine
  Learning (ICML)}, volume~37 of \emph{Proceedings of Machine Learning
  Research}, pages 1530--1538. PMLR, 2015.

\bibitem[Dinh et~al.(2017)Dinh, Sohl-Dickstein, and Bengio]{dinh2016density}
Laurent Dinh, Jascha Sohl-Dickstein, and Samy Bengio.
\newblock Density estimation using real {NVP}.
\newblock In \emph{International Conference on Learning Representations
  (ICLR)}, 2017.

\bibitem[Kingma and Dhariwal(2018)]{kingma2018glow}
Durk~P Kingma and Prafulla Dhariwal.
\newblock Glow: {G}enerative flow with invertible $1{\times}1$ convolutions.
\newblock In \emph{Advances in Neural Information Processing Systems 31
  (NeurIPS)}, pages 10215--10224. Curran Associates, Inc., 2018.

\bibitem[Poon and Domingos(2011)]{poon2011}
Hoifung Poon and Pedro~M. Domingos.
\newblock Sum-product networks: {A} new deep architecture.
\newblock In \emph{Conference on Uncertainty in Artificial Intelligence}, pages
  337--346, 2011.

\bibitem[Peharz et~al.(2020)Peharz, Lang, Vergari, Stelzner, Molina, Trapp, den
  Broeck, Kersting, and Ghahramani]{peharz2020}
Robert Peharz, Steven Lang, Antonio Vergari, Karl Stelzner, Alejandro Molina,
  Martin Trapp, Guy~Van den Broeck, Kristian Kersting, and Zoubin Ghahramani.
\newblock Einsum networks: Fast and scalable learning of tractable
  probabilistic circuits.
\newblock In \emph{International Conference on Machine Learning}, 2020.

\bibitem[Pevny et~al.(2020)Pevny, Smidl, Trapp, Polacek, and
  Oberhuber]{pevny2020SPTNs}
Tomas Pevny, Vasek Smidl, Martin Trapp, Ondrej Polacek, and Tomas Oberhuber.
\newblock Sum-product-transform networks: Exploiting symmetries using
  invertible transformations.
\newblock 2020.

\bibitem[Kingma and Ba(2015)]{kingma2014adam}
Diederik~P Kingma and Jimmy Ba.
\newblock Adam: {A} method for stochastic optimization.
\newblock In \emph{International Conference on Learning Representations
  (ICLR)}, 2015.

\bibitem[Bengio et~al.(2009)Bengio, Louradour, Collobert, and
  Weston]{bengio2009curriculum}
Yoshua Bengio, J{\'e}r{\^o}me Louradour, Ronan Collobert, and Jason Weston.
\newblock Curriculum learning.
\newblock In \emph{Proceedings of the 26th annual international conference on
  machine learning}, pages 41--48, 2009.

\bibitem[Grochow et~al.(2004)Grochow, Martin, Hertzmann, and
  Popovi{\'c}]{grochow2004style}
Keith Grochow, Steven~L Martin, Aaron Hertzmann, and Zoran Popovi{\'c}.
\newblock Style-based inverse kinematics.
\newblock \emph{ACM Transactions on Graphics}, 23\penalty0 (3):\penalty0
  522--531, 2004.

\bibitem[Aristidou and Lasenby(2009)]{aristidou2009inverse}
Andreas Aristidou and Joan Lasenby.
\newblock Inverse kinematics: {A} review of existing techniques and
  introduction of a new fast iterative solver.
\newblock Technical report, 2009.
\newblock CUED/F-INFENG/TR-632, Department of Engineering, University of
  Cambridge.

\bibitem[K{\"o}Ker(2013)]{koker2013genetic}
Ra{\c{s}}It K{\"o}Ker.
\newblock A genetic algorithm approach to a neural-network-based inverse
  kinematics solution of robotic manipulators based on error minimization.
\newblock \emph{Information Sciences}, 222:\penalty0 528--543, 2013.

\bibitem[Wang and Chen(1991)]{wang1991combined}
L-CT Wang and Chih-Cheng Chen.
\newblock A combined optimization method for solving the inverse kinematics
  problems of mechanical manipulators.
\newblock \emph{IEEE Transactions on Robotics and Automation}, 7\penalty0
  (4):\penalty0 489--499, 1991.

\bibitem[Kenwright(2012)]{kenwright2012inverse}
Ben Kenwright.
\newblock Inverse kinematics--cyclic coordinate descent (ccd).
\newblock \emph{Journal of Graphics Tools}, 16\penalty0 (4):\penalty0 177--217,
  2012.

\bibitem[Harvey et~al.(2020)Harvey, Yurick, Nowrouzezahrai, and
  Pal]{harvey2020robust}
F{\'e}lix~G Harvey, Mike Yurick, Derek Nowrouzezahrai, and Christopher Pal.
\newblock Robust motion in-betweening.
\newblock \emph{ACM Transactions on Graphics (TOG)}, 39\penalty0 (4):\penalty0
  60--1, 2020.

\bibitem[Pavllo et~al.(2019)Pavllo, Feichtenhofer, Auli, and
  Grangier]{pavllo2019modeling}
Dario Pavllo, Christoph Feichtenhofer, Michael Auli, and David Grangier.
\newblock Modeling human motion with quaternion-based neural networks.
\newblock \emph{International Journal of Computer Vision}, pages 1--18, 2019.

\bibitem[Bergamin et~al.(2019)Bergamin, Clavet, Holden, and
  Forbes]{bergamin2019drecon}
Kevin Bergamin, Simon Clavet, Daniel Holden, and James~Richard Forbes.
\newblock Drecon: data-driven responsive control of physics-based characters.
\newblock \emph{ACM Transactions on Graphics (TOG)}, 38\penalty0 (6):\penalty0
  1--11, 2019.

\bibitem[Won et~al.(2020)Won, Gopinath, and Hodgins]{won2020scalable}
Jungdam Won, Deepak Gopinath, and Jessica Hodgins.
\newblock A scalable approach to control diverse behaviors for physically
  simulated characters.
\newblock \emph{ACM Transactions on Graphics (TOG)}, 39\penalty0 (4):\penalty0
  33--1, 2020.

\bibitem[Kynk{\"a}{\"a}nniemi et~al.(2019)Kynk{\"a}{\"a}nniemi, Karras, Laine,
  Lehtinen, and Aila]{kynkaanniemi2019improved}
Tuomas Kynk{\"a}{\"a}nniemi, Tero Karras, Samuli Laine, Jaakko Lehtinen, and
  Timo Aila.
\newblock Improved precision and recall metric for assessing generative models.
\newblock In \emph{Advances in Neural Information Processing Systems 32
  (NeurIPS)}, pages 3929--3938. Curran Associates, Inc., 2019.

\bibitem[Ge et~al.(2015)Ge, McCool, Sanderson, and Corke]{ge2015modelling}
ZongYuan Ge, Chris McCool, Conrad Sanderson, and Peter Corke.
\newblock Modelling local deep convolutional neural network features to improve
  fine-grained image classification.
\newblock In \emph{IEEE International Conference on Image Processing (ICIP)},
  pages 4112--4116. IEEE, 2015.

\end{thebibliography}
\endgroup

\clearpage\appendix
\setcounter{section}{0}
\nipstitle{{Supplementary Material for} \\ Deep Residual Mixture Models}
\pagestyle{empty}


\section{Marginalisation in DRMMs}
\label{app:marginalisation}

Recall that the probability density function of a DRMM can be written as follows:
\begin{align}
    p(X^{(1)} \mid \vtheta) &= \sum_{h^{(1)}} \ldots \sum_{h^{(L-1)}} \, p^{(L)}(X^{(L)} \mid \vtheta) \nonumber \\
    & = \sum_{h^{(1)}} \ldots \sum_{h^{(L)}} w_{h^{(L)}} \, \prod^S_{s=1} f_{s}(\vx_s^{(L)} \mid h^{(L)}, \vtheta) \, .
    \label{eq:app:lh}
\end{align}

Let the first stream encompass all real-valued inputs. This allows us to express the above as:

\begin{equation}
p(X^{(1)} \mid \vtheta) = \sum_{h^{(1)}} \ldots \sum_{h^{(L)}} w_{h^{(L)}} \, \prod^S_{s=2} f_{s}(\vx_s^{(L)} \mid h^{(L)}, \vtheta) \, \mathrm{N}(\vx_1^{(L)} | \vmu_{1,h^{(L)}},(\sigma_1^{(L)})^2\MI)
\end{equation}

Now, note that $\mathrm{N}(\vx | \vmu,\MC)=\mathrm{N}(\vx+\vb | \vmu+\vb,\MC)$ for any vector $\vb$, and the residual computation gives $\vx_1^{(L)} =\vx_1^{(1)}-\widehat{\vx}_1^{(1)}-\ldots-\widehat{\vx}_1^{(L-1)}$. Therefore, we can select $\vb=\widehat{\vx}_1^{(1)}+\ldots+\widehat{\vx}_1^{(L-1)}$ to express the density in terms of the observable 1\textsuperscript{st} layer inputs $\vx_1^{(1)}$: 

\begin{align}
p(X^{(1)} \mid \vtheta) &= \sum_{h^{(1)}} \ldots \sum_{h^{(L)}} w_{h^{(L)}} \, \prod^S_{s=2} f_{s}(\vx_s^{(L)} \mid h^{(L)}, \vtheta) \, \mathrm{N}(\vx_1^{(L)} + \vb | \vmu_{1,h^{(L)}} + \vb,(\sigma_1^{(L)})^2\MI) \nonumber \\
&= \sum_{h^{(1)}} \ldots \sum_{h^{(L)}} w_{h^{(L)}} \, \prod^S_{s=2} f_{s}(\vx_s^{(L)} \mid h^{(L)}, \vtheta) \, \mathrm{N}(\vx_1^{(1)} | \sum_{l=1}^L \widehat{\vx}_1^{(l)},(\sigma_1^{(L)})^2\MI)), \label{eq:inputDRMM}
\end{align}

where $\vmu_{1,h^{(L)}} + \vb=\sum_{l=1}^L \widehat{\vx}_1^{(l)}$ follows from the reconstruction of real-valued layer input equaling the selected component mean, $\widehat{\vx}_1^{(l)}=\vmu_{1,h^{(l)}}$.

%

To allow marginalization, let $\vx_1^{(1)} = \{\vy_1^{(1)}, \vz_1^{(1)}\}$ denote the decomposition of $\vx_1^{(1)}$ into observed variables $\vy_1^{(1)}$ and those that we aim to marginalise out $\vz_1^{(1)}$. Then, by exchanging the order of integration and summation, the marginal probability density function of a DRMM is given by:
\begin{align}
    p(Y^{(1)} \mid \vtheta) &= \int_{\mbf{z}} \sum_{h^{(1)}} \ldots \sum_{h^{(L)}} w_{h^{(L)}} \, \prod^S_{s=2} f_{s}(\vx_s^{(L)} \mid h^{(L)}, \vtheta) \, \mathrm{N}(\vx_1^{(1)} | \sum_{l=1}^L \widehat{\vx}_1^{(l)},(\sigma_1^{(L)})^2\MI)) \, \dd \mbf{z} \nonumber \\
    &= \sum_{h^{(1)}} \ldots \sum_{h^{(L)}} w_{h^{(L)}} \, \prod^S_{s=2} f_{s}(\vx_s^{(L)} \mid h^{(L)}, \vtheta) \, \int_{\mbf{z}} \mathrm{N}(\vx_1^{(1)} | \sum_{l=1}^L \widehat{\vx}_1^{(l)},(\sigma_1^{(L)})^2\MI)) \, \dd \mbf{z} \nonumber \\
    &= \sum_{h^{(1)}} \ldots \sum_{h^{(L)}} w_{h^{(L)}} \, \prod^S_{s=2} f_{s}(\vx_s^{(L)} \mid h^{(L)}, \vtheta) \, \int_{\mbf{z}} \mathrm{N}(\vz_1^{(1)}, \vy_1^{(1)} | \sum_{l=1}^L \widehat{\vx}_1^{(l)},(\sigma_1^{(L)})^2\MI)) \, \dd \mbf{z} \nonumber \\
    &= \sum_{h^{(1)}} \ldots \sum_{h^{(L)}} w_{h^{(L)}} \, \prod^S_{s=2} f_{s}(\vx_s^{(L)} \mid h^{(L)}, \vtheta) \, \mathrm{N}(\vy_1^{(1)} | \sum_{l=1}^L \widehat{\vy}_1^{(l)},(\sigma_1^{(L)})^2\MI)) \, ,
\end{align}

In other words, the terms corresponding to the marginalized variables are simply omitted, which is in practice convenient to implement  using the multipliers $m_{r,n}$ in \cref{eq:multipliers}. Note that \cref{eq:multipliers} additionally considers the per-stream multipliers $m_c$ for categorical streams, assuming a general case where the first layer can have multiple input streams and types. Marginalizing over a categorical variable amounts to dropping out a whole stream, as each categorical stream represents a single variable. 

\section{DRMMs as GMMs}\label{app:inputspace}

\cref{sec:conditioning} informally observes that the sum of all layers' reconstructions $\sum_{l=1}^L \widehat{X}^{(l)}$ can be considered as an {\em input-space multilayer mixture component mean}. This can also be seen by collecting terms of \cref{eq:inputDRMM} as:
\begin{align}
    p(X^{(1)} \mid \vtheta) &= \sum_{h^{(1)}=1}^K \ldots \sum_{h^{(L)}=1}^K \overbrace{w_{h^{(L)}} \, \prod^S_{s=2} f_{s}(\vx_s^{(L)} \mid h^{(L)}, \vtheta)}^{=\gamma_{h^{(L)}}} \, \mathrm{N}(\vx_1^{(1)} | \sum_{l=1}^L \widehat{\vx}_1^{(l)}, (\sigma_1^{(L)})^2\MI)   \, \nonumber \\
&= \sum_{h^{(1)}=1}^K \ldots \sum_{h^{(L)}=1}^K \gamma_{h^{(L)}} \, \mathrm{N}(\vx_1^{(1)} | \sum_{l=1}^L \widehat{\vx}_1^{(l)},(\sigma^{(L)})^2\MI) \, .
\end{align}

This is in the form of a GMM with $K^L$ components resulting from all the possible additive combinations of the layer component means $\sum_{l=1}^L \widehat{\vx}_1^{(l)}=\sum_{l=1}^L \vmu_{1,h^{(l)}}$. The \textit{multilayer component weights} $\gamma_{h^{(L)}}$ are computed as a function of the last layer's component weights $w_{h^{(L)}}$ and the categorical stream $f_s$ terms. The latter, in turn, incorporate the other layer latents $h^{(1)},\ldots,^{(L-1)}$ through $\vx_s^{(L)}=\vx_s^{(1)}-\widehat{\vx}_s^{(1)}-\ldots-\widehat{\vx}_s^{(L-1)}=\vx_s^{(1)}-\vp_{s,h^{(1)}}-\ldots-\vp_{s,h^{(L-1)}}$.

\section{Implementation details}
\label{app:details}

\subsection{Animation Synthesis}
\paragraph{Data selection} The animation synthesis of \cref{sec:animation} utilizes the following Ubisoft LaForge dataset motion clips: {\em run1\_subject2,
run1\_subject5,
run2\_subject1,
run2\_subject4,
sprint1\_subject2,
walk1\_subject1,
walk1\_subject2,
walk1\_subject5,
walk2\_subject1,
walk2\_subject3,
walk2\_subject4,
walk3\_subject1,
walk3\_subject2,
walk3\_subject3,
walk3\_subject4,
walk3\_subject5.
}

The whole LaForge dataset contains highly diverse motions. The selected subset comprises all the walking and running locomotion clips with one sprinting clip omitted due to very extensive foot sliding in rapid turns. If the clip is included, it provides an easy ``shortcut'' and almost every synthesized direction change utilizes the sliding instead of performing more complex inference about foot placement. 

\begin{figure}[h]
  \centering
  \includegraphics[width=0.5\columnwidth]{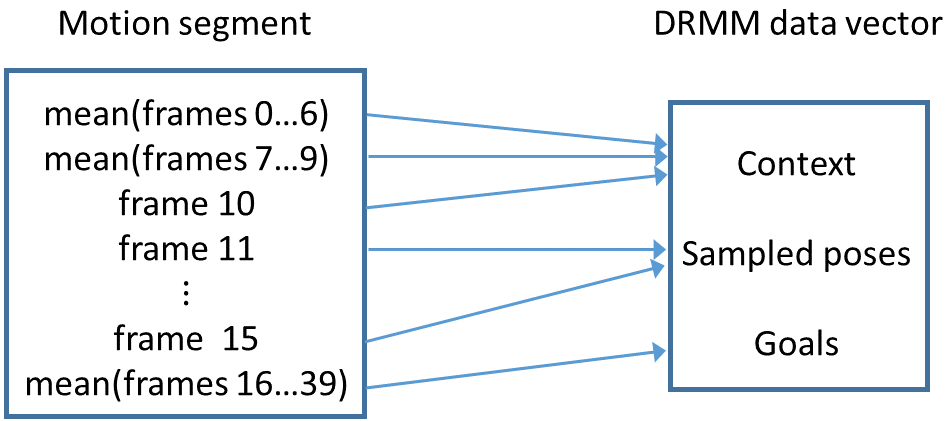}
  \caption{Partitioning of motion capture segments into sampled next poses and context and goals that the samples are conditioned on. \label{fig:mocap_partition}}
\end{figure}

\paragraph{Data preprocessing} The dataset was first imported to the Unity 3D game engine and global 3D joint positions (60 variables per frame) were exported from there as .csv files at 30 frames per second. This was split to segments of 40 frames, yielding a total of 152993 segments. The segments were partitioned as shown in \cref{fig:mocap_partition} into sampled next poses and the context and goals that the samples were conditioned on. The segments were also normalized by applying a translation and rotation such that in the current frame (last context frame), the character is at origin and facing the x-axis. 

The averaging of frames over time in the context and goals, shown in \cref{fig:mocap_partition}, reduces variance that is not relevant to the movement control task. For the goal frames, only the character root position was used and the average forward vector was also added to the goal variables. This allows conditioning independently on where the character should be facing and where it should move.  The resulting final training data vectors comprise a total of 487 variables each, \ie, this example demonstrates DRMM in a moderately high-dimensional problem. 

\paragraph{Sampling} During inference, the randomly placed flagpole's position gives the root position goal, and the direction of the goal gives the desired forward vector or facing direction. When only conditioning on movement target but not the facing direction, the forward vector is considered as unknown. As the goals can be beyond reach for some movement states (\eg, flagpole too far to reach within the limited planning horizon defined by the training data segment lengths), we utilize DRMM's flexible conditioning by first sampling a population of possible goals conditioned on the context. The mean and standard deviation of the samples is  used to clip the actual goals to within 3.5 standard deviations of the mean. The clipped goals are then used to condition the sampled next poses/frames. We sample and display 5 poses at a time. The sampled poses are appended to the end of the context to autoregressively inform subsequent samples.

\paragraph{Movement style} The data includes the actors switching between multiple locomotion styles like a crouched old man and happy hopping, which explains the occasional style change in the synthesized results, as we do not explicitly condition on style. Style conditioning should be possible, though, by labeling the motion frames into categories and using this data as an additional categorical input stream. To minimize manual style annotation labour, training should also be possible in a semi-supervised manner, by considering style as unknown for non-labeled frames.

\subsection{Truncated Sampling}\label{sec:truncation}
To reduce outliers, one can sample the latent $h$ with truncation, zeroing out component sampling probabilities below $\alpha_t \max_k p_k$, where $p_k$ is the sampling probability of mixture component $k$. We use $\alpha_t = 0.05$ during inference and $\alpha_t = 0.5$ during training. The higher truncation during training results in underestimating the variance of data modeled by each component. Although this allows a form of overfitting, it also improves mode separation in deep models. For instance, with $\alpha_t = 0.5$ during training, the 5-layer Sierpinsky likelihood in \cref{fig:sierpinsky} correctly shows all the component means, but $\alpha_t = 0.05$ makes the 5-layer likelihood blurred and approximately similar to the 4-layer model.

\subsection{Training Time}
We use 50k training iterations for the 2D plots, 500k iterations for IK and animation results, and 50k for the quantitative results in \cref{fig:benchmark}. Even the heaviest models can be trained in a matter of hours on a single NVIDIA 2080GTX GPU.

\subsection{Hyperparameters}
No exhaustive search of hyperparameters was conducted. Hyperparameters were iterated manually over a few months of development. Most parameters were initially decided based on testing with simple 2D data (\cref{fig:sierpinsky}), with which training a DRMM took less than a minute on a personal computer, allowing rapid iteration. \cref{table:hyperparameters} summarizes the hyperparameter choices.

\begin{table}[h]  \caption{Hyperparameters}\label{table:hyperparameters}
\centering
\begin{small}
\begin{tabular}{lll}
\toprule
{\em Parameter}  & {\em Value}  \\
\midrule
Learning rate & 0.005 (2D tests) \\
&  0.002 (other)\\
Laplace smoothing amount (\cref{sec:methods})  & $10^{-8}$ \\
Regularization loss weight (\cref{sec:regularization}) &  0.1  \\
Sampling truncation threshold  (\cref{sec:truncation})			 &  0.05 (inference) \\	
&  0.5 (training)\\
\bottomrule
\end{tabular}
\end{small}
\end{table}

\subsection{F1 Scores in \cref{fig:benchmark}}
The F1 score is the harmonic mean of precision and recall: $F_1=2 (\mathrm{recall} \times \mathrm{precision}) / (\mathrm{recall} + \mathrm{precision})$. We add $\epsilon$ to the denominator to handle zero precision and recall. We compute precision and recall using the method and code of \citet{kynkaanniemi2019improved}\footnote{\url{https://github.com/kynkaat/improved-precision-and-recall-metric}}, using 20k samples (or full dataset for smaller datasets) and batch size 10k. 

To save computing resources, we did not average the F1 scores in \cref{fig:benchmark} over multiple training runs. Nevertheless, the results should be reliable, as each plotted curve is already the result of multiple training runs, one per plotted point, and if there was significant randomness, the curves would not behave as consistently as in observed in \cref{fig:benchmark}.

\end{document}